\documentclass[sigconf]{aamas} 

\usepackage{balance} 
\usepackage{comment}
\usepackage{multirow}
\usepackage{enumitem}
\usepackage{algpseudocode}
\usepackage{algorithm}
\usepackage{ dsfont } 
\usepackage[flushleft]{threeparttable}
\usepackage{balance}

\newcommand{\states}{\mathcal{S}}
\newcommand{\stateatt}{S_t}
\newcommand{\stateattplus}{S_{t+1}}
\newcommand{\statezero}{S_0}
\newcommand{\stateone}{S_1}
\newcommand{\initialstates}{\rho_0}
\newcommand{\actions}{\mathcal{A}}
\newcommand{\actionatt}{A_t}
\newcommand{\actionzero}{A_0}
\newcommand{\actionone}{A_1}

\newcommand{\transitionfunc}{P}

\newcommand{\rewardfunc}{R}

\newcommand{\horizon}{T}
\newcommand{\policy}{\pi_{\theta}}
\newcommand{\positivereals}{\mathds{R}_{+}}
\newcommand{\trajectory}{\tau}
\usepackage{anyfontsize}         
\def\one{\mbox{1\hspace{-4.25pt}\fontsize{12}{14.4}\selectfont\textrm{1}}} 

\setcopyright{ifaamas}
\acmConference[AAMAS '21]{Proc.\@ of the 20th International Conference on Autonomous Agents and Multiagent Systems (AAMAS 2021)}{May 3--7, 2021}{Online}{U.~Endriss, A.~Now\'{e}, F.~Dignum, A.~Lomuscio (eds.)}
\copyrightyear{2021}
\acmYear{2021}
\acmDOI{}
\acmPrice{}
\acmISBN{}

\acmSubmissionID{610}

\title{Scalable Multiagent Driving Policies For Reducing Traffic Congestion}

\author{Jiaxun Cui}
\affiliation{
\institution{University of Texas at Austin}}
\email{cuijiaxun@utexas.edu}

\author{William Macke}
\affiliation{
\institution{University of Texas at Austin}}
\email{wmacke@cs.utexas.edu}
	
\author{Harel Yedidsion}
\affiliation{
\institution{University of Texas at Austin}}
\email{harel@cs.utexas.edu}

\author{Daniel Urieli}
\affiliation{
\institution{General Motors R\&D Labs}}
\email{daniel.urieli@gm.com}

\author{Peter Stone}
\affiliation{
\institution{University of Texas at Austin and Sony AI}}
\email{pstone@cs.utexas.edu}

\begin{abstract}
Traffic congestion is a major challenge in modern urban settings. The
industry-wide development of autonomous and automated vehicles (AVs) motivates
the question of how can AVs contribute to congestion reduction. 
Past research has shown that in small scale mixed traffic scenarios with
both AVs and human-driven vehicles, a small fraction of AVs executing a controlled multiagent driving policy can mitigate congestion. 
In this paper, we scale up existing approaches and develop new multiagent driving policies for AVs in scenarios with greater complexity. 
We start by showing that a congestion metric used by past research is manipulable in \emph{open road network} scenarios where vehicles dynamically join and leave the road. We then propose using a different metric that is robust to manipulation and reflects open network traffic efficiency.
Next, we propose a modular transfer reinforcement learning approach, and use it to scale up a multiagent driving policy to outperform human-like traffic and existing approaches in a simulated realistic scenario, which is an order of magnitude larger than past scenarios (hundreds instead of tens of vehicles).  
Additionally, our modular transfer learning approach  saves up to 80\% of the training time in our experiments, by focusing its data collection on key locations in the network.
Finally, we show for the first time a \emph{distributed}
multiagent policy that improves congestion over human-driven traffic. The distributed approach is more realistic and practical, as it relies solely on existing sensing and actuation capabilities, and does not require adding new  communication infrastructure.

\end{abstract}

\keywords{Autonomous Vehicles,
Deep Reinforcement Learning,
Traffic Optimization,
Multiagent Systems,
Multiagent Reinforcement Learning,
Flow}

\newcommand{\BibTeX}{\rm B\kern-.05em{\sc i\kern-.025em b}\kern-.08em\TeX}

\begin{document}


\pagestyle{fancy}
\fancyhead{}


\maketitle

\section{Introduction}
  Traffic congestion is one of the leading causes of lost productivity and
  decreased standard of living in urban settings~\cite{dresner2004multiagent}.
  Real world transportation systems suffer from inefficiency, partly due to the
  tendency of self-interested drivers to maximize personal utility over social
  welfare, and the inherent randomness in human driving~\cite{youn2008price}.
  The industry-wide development of autonomous and automated vehicles (AVs)
  motivates the question of how could AVs contribute to congestion reduction. 
  
  Since AVs are controlled by predefined policies, they can be made to act selflessly and drive strategically to influence human-driver behaviors to reduce congestion,  and thus increase the social welfare. For instance, an AV may smoothly slow down when reaching an on-ramp to allow for another vehicle to merge, causing following vehicles to smoothly slow down as well, and thus preventing propagation of \emph{stop-and-go waves} of congestion caused by sharp decelerations. 
  Past research has shown that in human-driven traffic, a small fraction of AVs  executing a controlled multiagent
  driving policy can mitigate congestion in simplified simulated and real-world
  scenarios~\cite{stern2018dissipation,vinitsky2018benchmarks}. 
  The research focusing on simulated
  experiments used Berkeley's Flow framework~\cite{wu2017flow}, which
  combines the SUMO traffic simulator~\cite{krajzewicz2012recent} with the RLlib
  deep reinforcement learning library~\cite{duan2016benchmarking}.  The simulated
  scenarios included cyclic road networks with a fixed set of vehicles, and
  more realistic non-cyclic road networks with vehicles joining and leaving,
  referred to as \emph{closed road networks} and \emph{open road networks} respectively.
  In this paper, we use the Flow framework to scale up existing approaches, and develop new multiagent policies for open road network scenarios with increased realism and complexity. 

  Past simulated open road networks were relatively small: a few hundreds of meters long, with a few tens of vehicles. Their small size allowed for using a centralized multiagent driving policy, which is limited in its ability to scale up
  since the observation and action spaces increase exponentially with the number
  of AVs. Moreover, we observed that the metric used by past research to show congestion improvement was manipulable by an RL agent in open road networks.
  The contributions of this paper are as follows:
  \begin{itemize}[leftmargin=*]
  \setlength{\itemsep}{0pt}%
    \setlength{\parskip}{0pt}%
    \item We outline the drawbacks of the time-average sample-average speed  metric used by past research (subsequently referred to as average speed metric),
and show empirically 
that in simulated open road networks
this metric is manipulable by a reinforcement learning (RL) agent.

\item We propose to use instead the \emph{outflow} congestion metric (rate of vehicles exiting the network), which reflects system level open network traffic efficiency. We highlight its advantages over the average-speed metric, and
show empirically that this metric is robust to manipulation.

    \item To avoid an exponential increase in complexity in a simulated realistic
    scenario that is an order of magnitude larger than in past research (hundreds
    instead of tens of vehicles, Figure~\ref{fig:i696}), we propose two RL approaches: (i) a modular approach where a 
    centralized policy is learned and operated locally around a key
    location in the larger network, and (ii) a transfer reinforcement learning approach where a policy learned in a small scenario is transferred to operate in a key location in the large scenario. Both of these multiagent policies outperform human-driven traffic and existing approaches.  
    
 \item We show for the first time a \emph{fully distributed} multiagent driving policy (using no communication) that improves congestion over human-driven traffic in an open road network scenario. Our distributed policy is more realistic and practical than a centralized one since (1) the size of the state and action space is independent of the number of vehicles, and (2) it relies solely on existing sensing and actuation capabilities, not requiring adding any new communication infrastructure. 
  \end{itemize}

The rest of the paper is organized as follows. Section \ref{sec:related_work} surveys related work. 
Section \ref{sec:background} defines the problem and the notation used. In Section \ref{sec:methodology} we compare evaluation metrics and outline our proposed solution methods. Section \ref{sec:empirical_evaluation} provides an empirical evaluation, the results of which are discussed in Section \ref{sec:results}. Section \ref{sec:conclusions} concludes and suggests future work.

\section{Related Work}
\label{sec:related_work}
  Traffic congestion has long been an active research area~\cite{downs2000stuck}.
  A common form of traffic jam in freeways is \emph{stop-and-go waves}, which were
  shown in field experiments to emerge when density exceeds a critical value,
  even with no apparent bottleneck~\cite{Sugiyama_2008}.  In recent field
  experiments, hand-designed controllers dissipated such waves and improved
  traffic flow~\cite{stern2018dissipation}. 
  
  The recent industry-wide development of autonomous and automated vehicles (AVs)
  has led to a surge of interest in harnessing AVs to reduce
  traffic congestion.
  On the theoretical side, there have been efforts to formalize and
  analyze the foundations for AVs impacting traffic
  systems~\cite{wu2018theoretical}.  
  On the applied side, large-scale traffic
  simulators have been adopted into a newly developed experimental framework
  called Flow~\cite{wu2017flow,vinitsky2018benchmarks}, 
  which we use in this paper. Using Flow, past research showed that
  Reinforcement Learning (RL)~\cite{sutton2018reinforcement} can learn an
  effective centralized multiagent driving policy, which simultaneously senses
  and controls all AVs, and improves the average traffic speed over human-driven
  traffic, implemented with accepted human driving
  models~\cite{Treiber2012TrafficFD,treiber2017intelligent}. However, we show
  that the average-speed metric is manipulable by an RL agent and might not
  accurately reflect the networks' traffic efficiency. Instead, we  propose using an
  alternative metric. 
  
  Since using RL to learn controllers in realistic simulated or real-world
  setups could be impractically slow, some research looked at using transfer
  learning~\cite{taylor2009transfer} to expedite learning, by transferring from
  a simulated ring to a simulated simple merge
  scenario~\cite{kreidieh2018dissipating}, and from a simulated to a scaled
  city~\cite{jang2019simulation}. Our transfer learning approach is different
  in the modular way it reuses state representation, which makes it more
  scalable. In the ring-to-merge transfer, the authors handled the different source and target scenario structure and size by assuming a maximum number of AVs in the
  road network, duplicating the ring state representation by this number, and
  using 0-padding if the actual number of AVs was smaller. In contrast, in our transfer
  approach the policy does not directly control more AVs than it was trained for. Instead,
  it is deployed only in a specific key location in the scenario, 
  and its state representation  remains the same even though the scenario has
  a different geometry and larger number of participants. In addition, the policy 
  transferred from ring to merge did not surpass the performance of a
  policy trained from scratch, while using our approach the transferred policy did. 
  
\begin{figure}
    \centering
    \includegraphics[width=\linewidth]{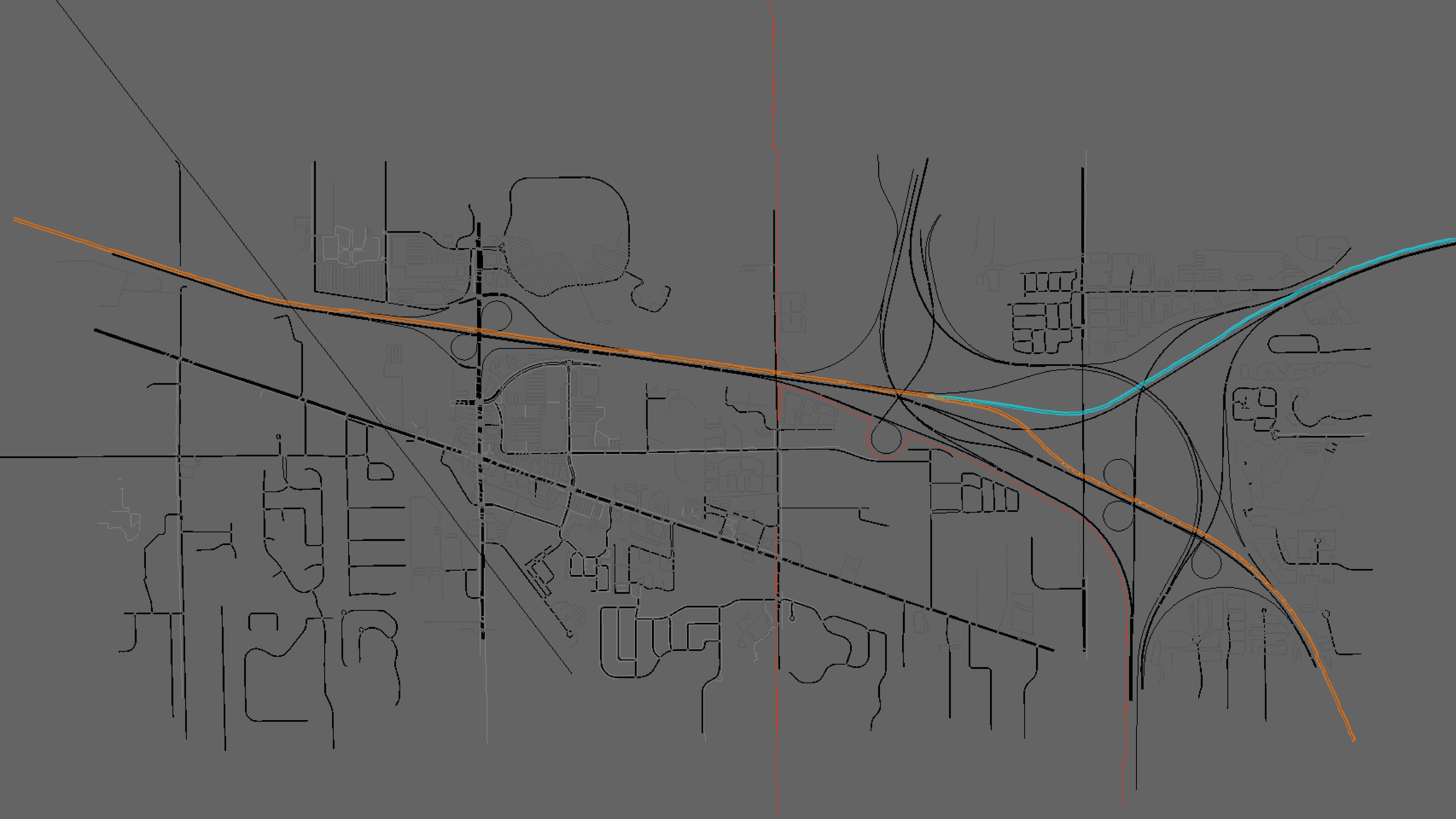}
    \caption{An image of highway \text{I-696} in Michigan, USA,  downloaded from
    Open Street Maps~\cite{OpenStreetMap}, as displayed in SUMO. It has a main road (colored orange) and a merging road (colored light blue). It is used in our experiments as a representative large-scale, realistic, open network.}
    \label{fig:i696}
  \end{figure}

  In the transfer learning from simulation to a scaled city,
  the source and target scenarios were the
  same and the state and action spaces were identical, so the challenge was not to
  generalize to a different scenario, but instead to compensate 
  for the differences between the simulation and the
  real world. Using our approach, we were able to scale to a simulated scenario 
  with hundreds of vehicles, an order of 
  magnitude larger than before, specifically  road \text{I-696} in
  Michigan, displayed in Figure \ref{fig:i696}, infamous for its high
  congestion. 

  To the best of our knowledge, this paper is the first to show a
  \emph{fully distributed} multiagent policy (using no communication) 
  that improves congestion over human-driven
  traffic in an open road network scenario. We are aware of two research papers 
  focusing on distributed
  multiagent policies for congestion reduction. The first relied on vehicle-to-vehicle 
  communication (V2V)~\cite{ibrahim2019control}, while we do not as it is currently
  unclear if and when V2V will be implemented in the real world.
  The second assumed that each distributed
  policy component is responsible for a \emph{region}, senses \emph{traffic images}, and
  uses \emph{communication infrastructure} to send actions composed of desired
  \emph{speed and headway}~\cite{maske2019distributedimages}. In contrast, each
  of our distributed policy components controls \emph{a single vehicle}, 
  senses its neighboring vehicles' \emph{speed and distance},
  uses \emph{no communication}, and sends direct \emph{acceleration actions}, all
  of which are motivated by maintaining high fidelity to real-world sensing and
  actuation capabilities. 

\section{Domain Description and Notation}
\label{sec:background}
We start by defining the traffic congestion reduction problem, its MDP formulation, and the traffic simulation environment we use.
\paragraph{Traffic congestion reduction problem definition.} Given an open road network (as defined in the introduction) with mixed autonomy traffic consisting of both human-driven vehicles and AVs, maximize the network's traffic efficiency by controlling the AV accelerations. 
Traffic Efficiency is measured in terms of \textbf{outflow} -- the number of vehicles per hour exiting the network.
A solution to the congestion reduction problem is a multiagent driving policy which maps the AV states to acceleration actions. We make the following assumptions:
        (i) Agents (AVs) are altruistic and have a common goal of reducing system congestion
        and (ii) Human drivers are self-interested and try to improve their own travel time.

\paragraph{MDP Definition}
The congestion reduction problems we address in this paper can be modeled as a discrete-time, finite-horizon Markov Decision Process (MDP)~\cite{puterman2014markov}, which is a tuple  $M = (\states, \actions, \transitionfunc,$ $\rewardfunc, \initialstates,\horizon  )$, where $\states$ is a state set, $\actions$ an action set, $\transitionfunc : \states \times \actions \times \states \rightarrow \positivereals$ a transition probability distribution, $\rewardfunc : \states \times \actions \rightarrow \mathds{R}$ a reward function, $\initialstates : \states \rightarrow [0,1]$ an initial state distribution, and T the time horizon. A \emph{driving policy} is a probability distribution  $\policy : \states \times \actions \rightarrow [0,1]$ parameterized by $\theta$ that stochastically maps states to driving actions. 

To find a solution policy, we train an RL agent to optimize a driving policy to maximize the expected return $E_{\trajectory} \left[ \sum_{t=0}^T \rewardfunc \left( 
\stateatt, \actionatt \right) \right]$, where $\trajectory := \left (\statezero, \actionzero, \stateone, \actionone \dots \right)$  denotes a trajectory, $\statezero \sim \initialstates$, $\actionatt \sim \policy \left (\cdot | \stateatt \right)$, $ \stateattplus \sim \transitionfunc \left( \cdot | \stateatt, \actionatt \right)$. In this paper, $\states$ is a set of AV observations, $\actions$ a set of acceleration actions, $\transitionfunc$ is computed by the simulator, and $\rewardfunc$ denotes the reward function. We discuss several implementations for the reward function in Section \ref{sec:evaluation_metrics}

\paragraph{Simulation Environment} 

We interface to the SUMO traffic simulator~\cite{krajzewicz2012recent}  using UC Berkeley's Flow software~\cite{kheterpal2018flow}. Flow provides OpenAI Gym~\cite{openai_gym} environments as wrappers around SUMO for easy interaction with various RL algorithm implementations. The simulator takes in maps of road structures, and simulates vehicle movements using accepted human driving
  models~\cite{Treiber2012TrafficFD,treiber2017intelligent} and definitions of \emph{inflows}, i.e., the location and rate of vehicles entering the network. The simulated vehicles follow safety and acceleration limits enforced by the simulator. A vehicle's \emph{leader} and \emph{follower} are the closest vehicles in front of and behind it (if exist). We note that 
  the actual inflow rate frequently differs from the requested one, for instance in cases where vehicles cannot enter the road network due to congestion. This opens an option for a vehicle to moderate the inflow by slowing down intentionally immediately after joining the network.
  The inability to guarantee exact inflows is the reason that the average-speed-based metric is not a valid congestion measure in open networks, as discussed in Section~\ref{sec:evaluation_metrics}.

 \section{Methodology}
\label{sec:methodology}

In this section we describe the evaluation metrics we use, and the structure of the centralized and distributed multiagent driving policies we train using RL. 
We discuss in detail the considerations that go into choosing appropriate \textit{Metrics} and \textit{Rewards}. Metrics are used for measuring the performance of a given policy, but are not always effective as RL rewards. In such cases, rewards different from the metric may be used as a performance measure for the RL agents.
\subsection{Evaluation Metrics}
\label{sec:evaluation_metrics}

In the Flow benchmark~\cite{vinitsky2018benchmarks}, the performance of the system is evaluated using \emph{time-average sample-average speed} over the episode, defined by Equation \ref{eq:avg_speed} 
\begin{equation}
    \text{Time-Average Sample-Average Speed}  \triangleq \frac{\sum_{t=1}^{T}\sum_{i=1}^{n_t} v_{i,t}/n_t}{T}
    \label{eq:avg_speed}
\end{equation} 
where $n_t$ is a time-dependant variable representing the number of vehicles in the traffic network at time t, $v_{i,t}$ is the instantaneous speed of vehicle $i$ at time $t$, and T is the episode length. 

In an ideal  
scenario with constant inflows, there are multiple metrics
that would all lead to the same ordering of policies: maximizing average speed, maximizing network outflow, and minimizing average time delay~\cite{kurt_AIM}.
In open road networks, a good policy should optimize the network outflow by maximizing the number of vehicles that pass through the network in a fixed time interval, however
policies that achieve high average speed might do so through manipulations that reduce inflows and outflows. 
For instance, one way to manipulate the average-speed metric is to block the incoming vehicles from entering the network until there is enough space for existing vehicles to accelerate to the maximum speed, thus maximizing the average-speed metric by compromising inflow and outflow.
 The average-speed metric is vulnerable because it ignores the (unmeasured) speeds of vehicles that haven’t entered the simulated road network. The outflow metric on the other hand is robust to this form of manipulation since delaying vehicles from entering the network is eventually penalized through reduced outflow.
 Therefore, we propose \emph{Outflow} as a performance metric in open networks as defined in Equation \ref{eq:outflow_metric}
\begin{equation}
    \text{Outflow} \triangleq \frac{\sum_t^{T}O_t}{T}
    \label{eq:outflow_metric}
\end{equation}
where T is the episode length and $O_t$ represents the number of vehicles that leave the network during timestep t. 

The reduction of inflows and outflows as a means of improving average speed is demonstrated in Table \ref{tab:t1}, which compares the results of using three reward functions --- the original Flow reward, 
the average-speed reward, and the outflow reward --- on Simple Merge defined in Section \ref{sec:reward_functions}.

\subsection{Centralized Multiagent Driving Policy}
\label{sec:centralized}
\begin{figure}
\centering
  \includegraphics[width=.90\linewidth]{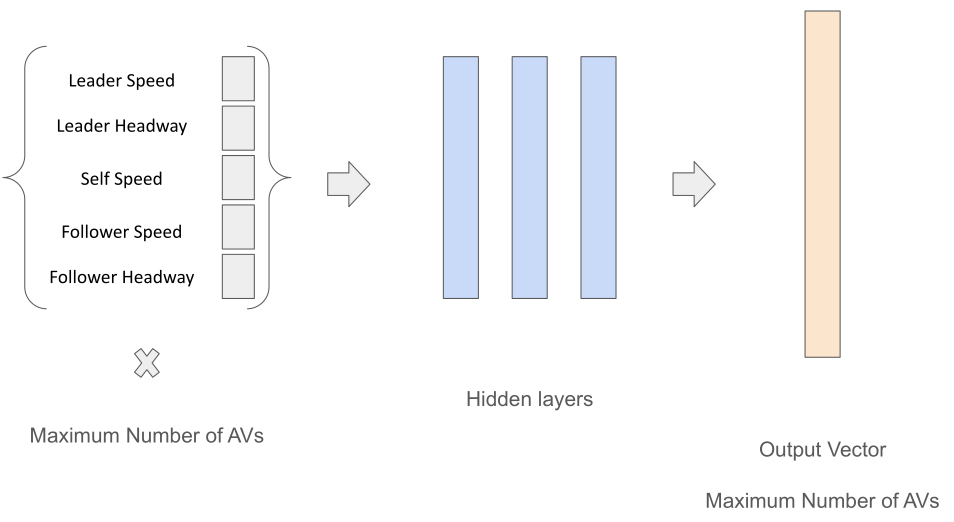}
  
  \caption[width=.9\linewidth]{Centralized neural network policy, where local states for vehicles are concatenated to form a global state. The state is passed through a series of hidden layers, resulting in an output vector of  accelerations of controlled AVs.
  \\
  }
        \label{fig:centralized}
  \centering
  \includegraphics[width=.95\linewidth]{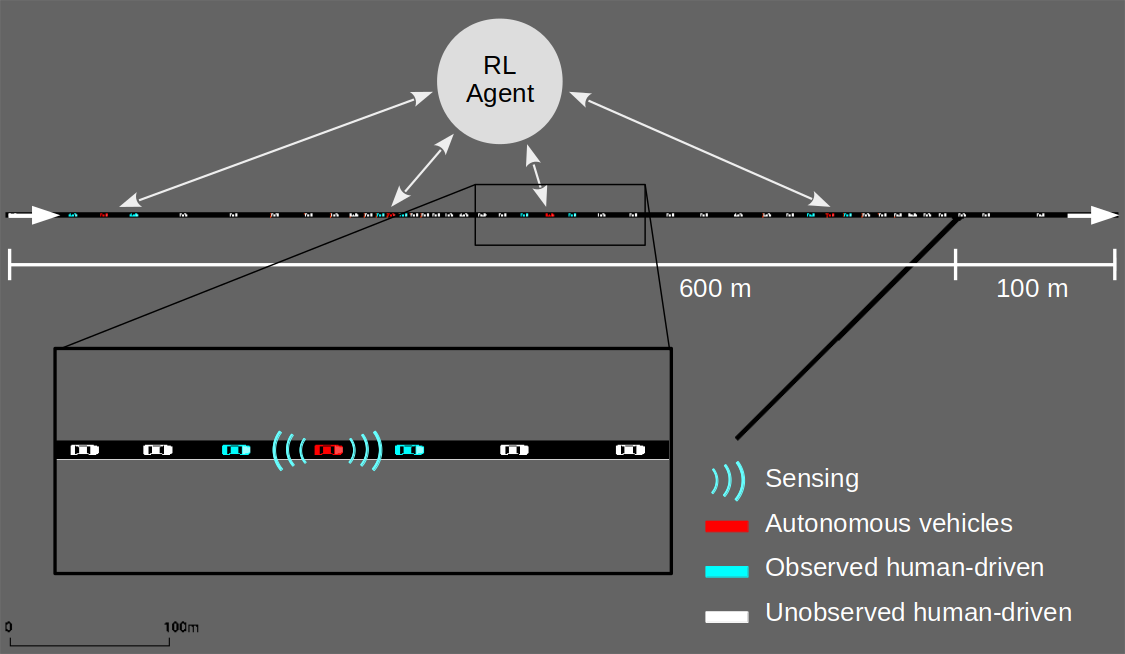}
  \caption[width=.9\linewidth]{Simple Merge network of length 700 m and
inflow rate 2000 veh/hr with an on-ramp of inflow rate 200 veh/h.
Perturbations caused by merging vehicles lead to stop-go waves congestion~\cite{kreidieh2018dissipating}.}
    \label{fig:merge}
\end{figure}

Our centralized driving policies are built on top of the ones used in previous work~\cite{vinitsky2018benchmarks,kreidieh2018dissipating}, 
where a centralized RL agent trained using the Proximal Policy Optimization (PPO) algorithm~\cite{Schulman2017Proximal}, controls a predefined fixed number of agents, $N_{AV}$ as illustrated in Figure~\ref{fig:centralized}. 
AVs are added to the list of controlled vehicles according to a FIFO rule based on when they entered the network. Below we discuss the state space and reward signal used for the centralized approach.

\subsubsection{State}
Similarly to past work~\cite{kreidieh2018dissipating}, at time-step $t$ the  centralized driving policy accepts a state observation $S_{t}$ which is  a concatenation of 5-tuples representing local AV states with the following values:
\begin{enumerate}
\setlength{\itemsep}{0pt}%
    \setlength{\parskip}{0pt}%
    \item Normalized speed of the $AV_i$, $v_i$
    \item Normalized speed of the leader of $AV_i$, $v^{L}_i$
    \item Normalized headway between $AV_i$ and its leader, $h^{L}_i$ 
    \item Normalized speed of the follower of $AV_i$, $v^{F}_i$
    \item Normalized headway between $AV_i$ and its follower, $h^{F}_i$
\end{enumerate}
The speed values are normalized by the max possible speed $V_{max}$, and the headway values are normalized by a constant representing the maximum possible headway, $h_{max}$.
Suppose the maximum number of AVs controlled by the centralized policy is $N_{AV}$, then the state feature $S_{t}$ is a vector of length 5$N_{AV}$, and is padded with zeros when the number of AVs in the network is smaller than $N_{AV}$. Formally, the state of $AV_i$ at time $t$, $S_{i,t}$ is defined in Equation \ref{eq:sti}, and the concatenated state of all the AVs at time $t$, $S_{t}$ is defined in Equation \ref{eq:st}.
\begin{equation}
    \begin{aligned}
    \label{eq:sti}
    S_{i,t} = [\frac{v_{i,t}}{V_{max}},\frac{v^{L}_{i,t}}{V_{max}},\frac{h^{L}_{i,t}}{h_{max}},
    \frac{v^{F}_{i,t}}{V_{max}},\frac{h^{F}_{i,t}}{h_{max}}]
    \end{aligned}
\end{equation}
\begin{equation}
    \begin{aligned}
        \label{eq:st}
    S_{t} = [S_1,S_2,....S_{N_{AV}}]
    \end{aligned}
\end{equation}

\subsubsection{Reward} 
\label{section:simple_merge_reward}
There are several possible objectives to optimize for in open networks, such as maximizing network outflow, or minimizing the maximum time delay of any vehicle to prevent starvation. In this paper, we focus on maximizing the efficiency of a network in the form of average outflows.
There are three reward functions considered in our experiments.
\paragraph{\textbf{Original Flow Reward~\cite{vinitsky2018benchmarks}}}
The reward in the Flow benchmark is composed of $\ell_2$-norm distance to a desired velocity and a small-headway penalization term. This reward encourages every vehicle to travel as close as it can to the desired speed every time step while maintaining a large headway.
\begin{equation}
    r_t = \max(\Vert V_{d} {\mathbf{1}}^{n}\Vert_2 -\Vert V_{d}-v\Vert_2,0)/\Vert V_{d} {\mathbf{1}}^{n}\Vert_2 - \alpha \sum_{i\in AV}\max(h_{max}-h_i,0)
\end{equation}
where $v$ is a speed vector of all the vehicles in the network, $V_d$ is the desired speed scalar, $\mathbf{1}^n$ is a $\mathbf{1}$ vector with n elements, where n is the total number of vehicles in the network, $\alpha$ is an adjustable constant, $h_{i}$ is the headway between the $i$th AV and its leader, and $h_{max}$ is a constant of expected headway. 
\paragraph{\textbf{Average Speed Reward}}
We define an instantaneous average speed reward as
\begin{equation}
    r_t = \frac{\sum_{i=1}^{n} v_i}{nV_{max}}
\end{equation} 
where n is the current number of all vehicles in the traffic network, and $V_{max}$ is the maximum speed allowed on every lane. This reward is provided every time step. Summing it over the entire episode and then dividing the sum by the episode's horizon $T$ gives the value of average speed (Equation \ref{eq:avg_speed}) of the episode.
\paragraph{\textbf{Outflow Reward}}
The reward for instantaneous outflow is 
\begin{equation}
    r_t = O_t
\end{equation}
where $O_t$ is the number of vehicles that leave the observed area of the traffic network through any lane during the $t$th time step. We note that the sum of this reward over the simulation will always be proportional to the Outflow metric (Equation~\ref{eq:outflow_metric}) by a factor of $1/T$, assuming the simulation occurs over a fixed period of time. Thus optimizing this reward is equivalent to optimizing  the Outflow metric.

\subsection{Modular Transfer Learning Approach}
\label{sec:modular}

Scaling up to the \text{I-696} Merge scenario results in an order of magnitude more vehicles (hundreds instead of tens).
Training RL agents in this scenario is challenging for at least three reasons. First, the state and action 
space grow exponentially with the number of controlled vehicles when using a centralized approach. Specifically, the combined action space of the system is, $|A|=|A_i|^{N_{AV}}$ 
where $|A_i|$ is the size of the action space of a single AV, and the size of the combined state space is $|S|=|S_i|^{N_{AV}}$ where $|S_i|$  is the size of the state space of a single AV.
Second, while the centralized agent's most important actions are those that are near the merge point,  the congestion-related rewards (Section~\ref{section:simple_merge_reward}) are calculated based on all vehicles in the network, most of which are more impacted by their own actions than by the centralized controller, so that the agent's reward is very noisy.
Third, there is a large delay in rewards due to the delay in the effect of an AV action on the system's average speed and outflow. 
Therefore, in the I-696 scenario we use transfer reinforcement learning and a modular approach. 
\subsubsection{Method}
We create a window surrounding the junction so that the length of each road segment is comparable to a corresponding segment in a smaller network  we trained on. We then take a policy that was trained in the small network, and apply it to the AVs inside the window, while outside of the window AVs act like human drivers.
We refer to this approach as the \textit{Zero-Shot Transfer} approach,
and compare it to training from scratch within this same window, referred to here as \textit{Train from scratch (Window)}.
\subsubsection{State and Reward}
The states and rewards employed in the modular approach are the same as in the centralized method.

\subsection{Distributed Multiagent Driving Policy}
\begin{figure}
    \centering
    \includegraphics[width=\linewidth]{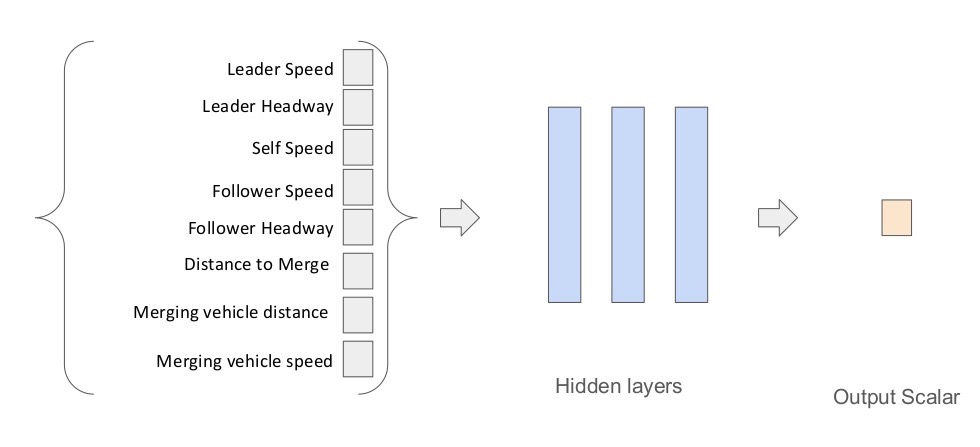}
    \caption{Distributed model, where each vehicle only has access to local observations. The local observation is passed through hidden layers, resulting in the final scalar output of the AV acceleration.  This same policy is applied to every AV in the network, each with its own local observations.}
    \label{fig:distributed}
\end{figure}

In our distributed setting, autonomous agents share the same policy which is  executed  locally as Figure~\ref{fig:distributed} shows. Each agent only has access to its local observations, and acts independently from other agents. In the training process, each agent receives its own reward, and  the experiences of all agents are used to train the same policy.

\subsubsection{State} 
In the distributed setting, AVs rely only on their own sensed information and lack the information of the entire network that the central policy has. To mitigate this lack of information, we include both the original state features for a single AV of the centralized method as well as several additional features which can be obtained using the AV's sensors, including: distance from agent to the next merging point; the speed of the next merging vehicle and its distance to the merge junction. With this added information the state becomes:

\begin{equation}
    \begin{aligned}
    S_{i,t} &= [\frac{v_{i,t}}{V_{max}},\frac{v^{L}_{i,t}}{V_{max}},\frac{h^{L}_{i,t}}{h_{max}},\frac{v^{F}_{i,t}}{V_{max}},\frac{h^{F}_{i,t}}{h_{max}}, \\
  &\frac{d_{next}}{d_{max}},\frac{v_{merge}}{V_{max}},\frac{d_{merge,next}}{d_{max}}]
    \end{aligned}
\end{equation}
where $d$ measures the length along the predefined route in $m$, 
$v$ denotes speed in $m/s$, $h$ denotes headway in $m$, $V_{max}$ is the max possible speed, $d_{max}$ is a constant that evaluates the length from the network entry to the merging junction, and $h_{max}$ is a constant representing the max possible headway. In particular, $v_i,v^{L}_i,v^{F}_i$ denotes speed of the $i$th AV at time step t, 
its nearest leader's speed, and its nearest follower's speed respectively; $h^{L}_i$ and $h^{F}_i$ denote headway between the $i$th AV and its leader, and the headway between the $i$th AV and its follower; $d_{next}$ represents the distance between the $i$th AV
and the next junction on its route, and $d_{merge,next}$ is the minimal distance of all vehicles on a different edges to the junction.
Since the policy is shared among all agents in the traffic network, the number of AVs can vary among different environments, and the theoretical maximum value that $i$ can take is the maximal number of  vehicles allowed in the observed traffic network.

\subsubsection{Reward}
\label{sec:distributed_reward}
The reward design in the distributed setting is different from that in the centralized setting, since the agent only gets rewards while it is in the simulation. The Outflow reward is only affected by the AV's actions after the AV had exited the simulation, so the agent does not get to observe its own rewards. 
The Average Speed reward does not encourage the agents to exit the simulation, since higher rewards (speeds) result in spending less time in the simulation and therefore lower cumulative rewards. As a result the agents may reduce their speed without incurring a substantial reduction in their return.
A possible alternative is to penalize an agent for every time step it stays in the network (i.e. for agent $i$, the reward at time $t$ would be $r_{i,t}=-0.1$). We refer to this as  \emph{selfish} reward.
We found in preliminary experiments that if agents are only rewarded according to individual time-delay, their learned policy is inferior to a policy trained using a  combination of selfish and collaborative reward in distributed shared policy training.

We use $\eta_1$ to denote the weighting of the individual time-delay penalty (selfish component), and $\eta_2$ to denote the weighting of the system average speed (collaborative component), where $\eta_1+\eta_2=1, \eta_1 \geq 0, \eta_2\geq 0$.

The effectiveness of mixing reward in this way is consistent with previous work on multiagent reward mixing~\cite{IJCAI20-ishand}. Our preliminary experiments also show that a bonus for each agent upon leaving the traffic network helps, so the final reward used in our distributed approach is defined as:
\begin{equation}
    \label{final_distributed_reward}
    r_{i,t}=(1-\one{done})(-\eta_1+\eta_2\times\frac{\sum_{j=1}^{n} v_j}{nV_{max}})
         + \one{done}\cdot Bonus
\end{equation}
Where $\one{done}$ is an indicator function that takes value 1 at the time the agent vehicle leaves the network, and 0 when the vehicle is still in the network.
We perform sensitivity analysis on the values of coefficients of the reward function in Table \ref{tab:distributed_reward}

\section{Empirical Evaluation}
In this section we describe the properties of the two types of road networks that we use in our empirical evaluation, the Simple Merge, and the I-696 highway. We also describe the characteristics of the two types of vehicles in the network, human driven vehicles, and AVs.
\label{sec:empirical_evaluation}

\subsection{Traffic Scenario 1 - The Simple Merge}
\label{sec:simple_merge}
Our Simple Merge experiments are based on the Flow 
benchmark~\cite{vinitsky2018benchmarks}.
The road network consists of a main highway of length 600m before the merge and a merging lane of 200m. After merging, the vehicles still need to travel an additional 100m. The junction controller is a "priority" controller where both incoming edges have equal priority. This controller is the same as in previous benchmarks~\cite{vinitsky2018benchmarks}. If two vehicles arrive at the junction at the same time with equal priority, the one with the lower speed will yield to the vehicle with the higher speed. The main highway has an inflow of 2000 vehicles per hour consisting of 90\% humans and 10\% AVs. The merging lane has an inflow of 200 vehicles per hour, made up entirely of human drivers.
This setup is compatible with real highway capacity levels of 2250 vehicles/hour/lane \cite{laufer2007freeway}.
For both inflows, vehicles enter the traffic network according to the predefined inflow rate with some small stochastic variance in the arrival times. In the mixed autonomy traffic flow, the AVs are equally spaced among human vehicles. In the centralized policy the maximum number of controlled AVs, $N_{AV}$, is 5. In the distributed policy there is no limit on the number of controlled AVs.

\subsection{Traffic Scenario 2 - The \text{I-696} Merge}
\label{sec:i696}
The \text{I-696} network has the same shape as the real-world Interstate 696 highway in the US, which is a much larger network than the simple merge.
 In our experiments we simplified the \text{I-696} network to have a single rather than multiple lanes, a main road, and a single merging road as highlighted 
 in Figure \ref{fig:i696}. We refer to this part of the network as the \text{I-696} Merge. The \text{I-696} Merge is much longer than the Simple Merge, which makes it challenging for existing methods to learn effective driving policies. 
The highway length before the merge is 3131m, the merging edge length is 1878.56m, and after merging the vehicle still needs to travel 5077.7m. The defined traffic inflows are the same as in the Simple Merge.

\subsection{Human-Driven Vehicles}
In all scenarios,  human-driven vehicles  are modelled using an \textit{Intelligent Driver Model} (IDM)~\cite{treiber2000congested} that tries to drive at a maximal speed while maintaining at least a 1-second gap from its leader .

\subsection{Autonomous Vehicles (AV)}
Autonomous vehicles are only included in the main highway inflow with a 10\% penetration rate and equal spacing. There are at most 5 controlled AVs in centralized Simple Merge, 30 in the centralized \text{I-696} Merge, and any number in the distributed Simple Merge. 

\subsection{Training Details}
All experiments are trained with the same set of parameters using the Proximate Policy Optimization (PPO) algorithm~\cite{Schulman2017Proximal}. Both tasks were trained in an episodic manner with a horizon of 2000 time steps of length 0.5 seconds. All results are obtained from SUMO 1.6.0 and Ray 1.0.1.
\footnote{More details on the training hyper-parameters and MDP parameters are provided in the appendix of the full version here: \url{https://www.cs.utexas.edu/~aim/papers/AAMAS2021.pdf}} \footnote{Our code base is publicly available here: \url{https://github.com/cuijiaxun/MITC-Project}}

\section{Results}
\label{sec:results}
In all our experiments, for each configuration we execute three policy learning runs, select the learned policy with the highest return, and evaluate its performance in 100 simulations using a fixed set of 100 random seeds (which affect the arrival times of the vehicles entering the network). We report the mean values of relevant metrics accompanied with their 95\% confidence interval error bounds.
\footnote{Videos can be found here: \url{ https://www.cs.utexas.edu/~aim/flow.html}}
\subsection{Comparison of Reward Functions}
\label{sec:reward_functions}
Table \ref{tab:t1} shows the results of training the centralized policy described in Section \ref{sec:centralized} in the Simple Merge scenario described in Section \ref{sec:simple_merge}, using each of the three reward functions described in Section \ref{section:simple_merge_reward} (along with human driven traffic as a baseline). 
\begin{table*}[h!]
  \centering
  \caption{Statistics of Reward Functions on Simple Merge}
    \label{tab:t1}
  \begin{threeparttable}
    \begin{tabular}{|c|c|c|c|}
     \hline $Reward$ & Average Outflow (vehs/hr)& Average Inflow (vehs/hr) & Average Speed (m/s) \\
     \hline Human &1558.12$\pm$2.99& 1725.48$\pm$2.89 & 7.27$\pm$ 0.15\\
     \hline Original Flow Reward &1724.55$\pm$6.98 &1769.36$\pm$6.60 &18.95$\pm$0.19\\
     \hline Average Speed Reward &1379.45$\pm$2.99 & 1408.46$\pm$3.28 & \textbf{19.34}$\pm$0.02\\ 
     \hline Outflow Reward&\textbf{1804.21}$\pm$7.17 & \textbf{1864.55}$\pm$7.24 & 16.21$\pm$0.08\\ 
     \hline
    \end{tabular}
    \end{threeparttable}
\end{table*}

All reward functions --- the original Flow reward, 
the average-speed reward, and the outflow reward --- result in  improved average speed over the human baseline, where the average speed reward results in the highest average speed in the network.
However, we see that this improvement comes at the cost of overall reduced network throughput, even when compared with the human baseline. The Average Speed reward produced network inflows and outflows that were significantly lower than the human baseline (an independent T-Test yields p-values $<0.001$ for both metrics). By contrast, both the Flow reward function and the outflow reward function 
are able to increase all 3 metrics compared to the human baseline, however the Outflow reward function still outperforms the Flow reward in terms of outflow and inflow 
by a statistically significant margin. An independent T-Test yields p-values $<0.001$ for both inflows and outflows.

\subsection{Modular Transfer Learning}
\label{sec:transfer}
In this section we compare the performance of three RL approaches and human-driven traffic on the \text{I-696} Merge scenario (Section \ref{sec:i696}):
\begin{itemize}[leftmargin=*]
    \setlength{\parskip}{0pt}%
\item  Zero-Shot Transfer approach (Section \ref{sec:modular})
\item  Train from scratch (Window) approach (Section \ref{sec:modular})
\item Train from scratch ($N_{AV}$=30) approach --- trained on the entire \text{I-696} Merge with a maximal number of controlled  AV,  $N_{AV}=30$, and applied to up to 30 AVs in \text{I-696} Merge.
\end{itemize}
All three approaches were trained using two reward functions:  the Flow reward, and  the Outflow reward. Table \ref{tab:3} demonstrates that the Zero-Shot Transfer approach integrated with the outflow reward produces the best outflow results: significantly better than human performance. The next best approach is the Train From Scratch in a window approach, in combination with the Outflow reward. However the difference between these top two approaches is not statistically significant with p-value=0.141 in an independent T-Test.
The top two approaches also have better average speed than the human baseline and comparable inflows. 
The Train from scratch ($N_{AV}$=30) approach performs worse than the human baseline under both reward functions, as it was unable to address the three challenges described in Section \ref{sec:modular}. 
\begin{table*}[h!]
  \centering
  \caption{Transferring a policy from Simple Merge to \text{I-696} Merge}
  \label{tab:3}
  \begin{threeparttable}
  \begin{tabular}{|c|c|c|c|c|}
     \hline Experiment & Reward& Average Outflow(vehs/hr) & Average Inflow(vehs/hr) & Average Speed (m/s)\\
     \hline Human &None&936.90$\pm$5.96& 2184.91$\pm$0.29 & 16.27$\pm$0.12\\
     \hline
     \multirow{2}{*}{ Train From Scratch ($N_{AV}$=30)}  
     & Outflow &366.98$\pm$1.91& 561.60$\pm$3.12& 19.57$\pm$0.27\\\cline{2-5}
     & Flow reward & 638.06$\pm$10.99& 1165.06$\pm$10.22 & 14.95$\pm$0.12\\\cline{2-5}
     \hline 
     \multirow{2}{*}{ Train From Scratch (Window) }  
     & Outflow & \textbf{1012.64}$\pm$9.23& 2178.76$\pm$2.81 & 16.99$\pm$0.13\\\cline{2-5}
     & Flow reward & 923.29$\pm$5.79& 2181.17$\pm$1.92 & 15.98$\pm$0.10\\\cline{2-5}
     \hline 
     \multirow{2}{*}{ Zero-Shot Transfer (Window)}  
     & Outflow &\textbf{1017.32}$\pm$10.49 & 2170.55$\pm$4.61 & 17.05$\pm$0.16 \\\cline{2-5}
     & Flow reward& 928.00$\pm$6.06& 2181.53$\pm$1.67 & 16.09$\pm$ 0.11\\\cline{2-5}
     \hline
    \end{tabular}
    \end{threeparttable}
\end{table*}
The original Flow reward never beats the human baseline in terms of outflows, in any of the training approaches.

Note that outflows in I-696 are approximately half of those in Simple Merge due to the length of I-696, since vehicles take a long time to reach the end of the simulated highway. 
Since the simulation on \text{I-696} is much slower than on Simple Merge, training on \text{I-696} takes approximately 5 times longer for the same number of iterations than training on Simple Merge.


\subsection{Distributed Setting}
In this section we perform feature selection on the distributed approach's state representation, and conduct sensitivity analysis on the hyper-parameters of the distributed reward function.
\begin{table*}[h!]
  \centering
  \caption{Statistics of Evaluation of Distributed Method Using Different Features}
  \label{tab:ablation}
  \begin{tabular}{|c|c|c|c|c|}
     \hline Augmentation & Episodic Return & Average Outflow(vehs/hr) & Average Inflow(vehs/hr) & Average Speed (m/s) \\
     \hline Human &Not Applicable &1558.12$\pm$2.99& 1725.48$\pm$2.89 & 7.27$\pm$ 0.15\\
     \hline No Augmentation &458.09$\pm$9.65 &1610.68$\pm$10.56& 1658.45$\pm$10.64 & 16.02$\pm$0.17\\
     \hline Full Augmentation&\textbf{476.81}$\pm$14.47 &\textbf{1791.07}$\pm$6.60 &1850.72$\pm$6.76 &15.91$\pm$0.05\\
     \hline Dist &447.64$\pm$13.26 &1663.49$\pm$10.95 & 1725.55$\pm$9.94& 14.57$\pm$0.31\\ 
     \hline Dist+MergeInfo &444.95$\pm$13.33 &1674.72$\pm$9.72 & 1741.90$\pm$7.76 & 14.40$\pm$0.22\\ 
     \hline MergeInfo &434.43$\pm$8.21 &1600.67$\pm$9.28 & 1657.01$\pm$9.34 & 15.04$\pm$0.14\\ 
     \hline Congestion+MergeInfo &456.05$\pm$13.18 &1666.55$\pm$14.97&1726.24$\pm$14.99 &14.92$\pm$0.22\\
     \hline Congestion+Dist &425.87$\pm$4.17 &1686.24$\pm$6.49&1755.50$\pm$7.39 &13.53$\pm$0.06\\
     \hline
\end{tabular}
\end{table*}

\subsubsection{Distributed State Feature Augmentation}
The centralized agent receives local observations sent from all autonomous vehicles, which can provide indirect information indicating the traffic situation at different locations over the network. For example, the speed of first AV may represent the congestion level ahead of the second AV, so the second AV can adjust its behavior according to this. In our distributed setting, however, no information is communicated between AVs, so the agents have to make decisions solely based on local information. 

One intuition is that AVs can make better decisions if they are provided with system-level information.
The following environmental information is hypothesized to be useful for the distributed agents to learn a policy that can improve traffic efficiency. For simplicity, we will use the abbreviations in parentheses for each feature for future reference. 
\begin{enumerate}
    \item Average speed of vehicles between the AV and the next junction (Congestion) 
    \item Distance from the AV to the next junction (Dist) 
    \item Distance from the first vehicle that is going to merge to the junction and the speed of this merging vehicle (MergeInfo)
\end{enumerate}

We show in Table \ref{tab:ablation} that if the agents totally rely on the speed and headway information of itself, its leader, and its follower without any extra information (referred to as ``No Augmentation''), they can achieve slightly better results than the human baseline, but additional state information further improves the performance.
The experiments were conducted using a ``0.1-Collaborative reward'' (the reward defined in Equation     \ref{final_distributed_reward} with $\eta_1=0.9$, $\eta_2=0.1$) and a bonus for completion ($r_{i,done}=+20$). We noticed that when including the congestion feature, the learning process became less stable, with average return decreasing later in the training process. We conjecture that this happens because the observed values of this feature are highly dependent on the AVs' current policy. For instance, during early training, the agents may experience mainly high congestion values, while once the policy improves, it sees mainly low congestion values.  Therefore including this feature adds an additional non-stationary element to the learning process.
Table \ref{tab:ablation} shows how state augmentation affects the training process. The model with the full set of state features (Dist, MergeInfo, and Congestion) performed the best.




\begin{table*}[h!]
  \centering
  \caption{Comparison of different reward function parameters of the Distributed Method on Simple Merge}

  \begin{tabular}{|c|c|c|c|}
     \hline $\eta_1,\eta_2,r_{i,done}$ & Average Outflow (vehs/hr) & Average Inflow (vehs/hr) & Average Speed (m/s) \\
     \hline Human &1558.12$\pm$2.99& 1725.48$\pm$2.89 & 7.27$\pm$ 0.15\\
     \hline $\eta_1=1,\eta_2=0,+0$ &1749.78$\pm$7.78 & 1807.99$\pm$7.93 & 16.01$\pm$0.06\\ 
     \hline $\eta_1=1,\eta_2=0,+20$ &1771.74$\pm$5.63 & 1831.75$\pm$5.98 & 15.91$\pm$0.04\\ 
     \hline $\eta_1=0.9,\eta_2=0.1,+20$ &\textbf{1791.07}$\pm$6.60 & 1850.72$\pm$6.76 & 15.91$\pm$0.05\\ 
     \hline $\eta_1=0.9,\eta_2=0.1,+0$ & 1622.34$\pm$6.74 & 1685.02$\pm$6.86 &13.99$\pm$0.06\\
     \hline $\eta_1=0.8,\eta_2=0.2,+20$ & \textbf{1796.76}$\pm$6.78 & 1856.70$\pm$7.07 &16.03$\pm$0.05\\ 
     \hline $\eta_1=0.7,\eta_2=0.3,+20$ &1740.64$\pm$5.14 & 1801.58$\pm$5.30 & 15.38$\pm$0.04\\ 
     \hline $\eta_1=0.5,\eta_2=0.5,+20$ &1750.46$\pm$6.51 & 1809.83$\pm$6.72& 15.59$\pm$0.23\\ 
     \hline $\eta_1=0,\eta_2=1,+20$ &1744.96$\pm$6.69&  1808.28$\pm$7.07 & 13.11$\pm$1.09\\
     \hline $\eta_1=0,\eta_2=1,+0$ &271.44$\pm$6.29 & 566.64$\pm$3.96 & 1.54$\pm$0.02\\ 
      
     \hline
     \end{tabular}
     \label{tab:distributed_reward}
\end{table*}
\subsubsection{Distributed Reward - Parameter tuning}
Next, we compare the resulting performance when learning policies with
the distributed reward defined in Equation \ref{final_distributed_reward}, parameterized with
different values of $\eta_1$ (selfish), $\eta_2$ (collaborative) and $Bonus$ (completion).

In Table \ref{tab:distributed_reward} we can see that when the reward is purely collaborative  ($\eta_1=0$, $\eta_2=1$,$+0$), it does not encourage agents to leave the system, since the outflow of $271.44$ is about $17\%$ of the human baseline outflow. In fact, the longer an agent stays in the system, the more reward it can get at the early training stage. As a result, the policy optimization can become trapped at a local optimum. Visualization of the resulting policy  shows that at some point an AV stops and lets the merging vehicles travel at full speed so as to  gain more speed-based reward.

We see that by either moving to a fully selfish reward, or adding an exit bonus reward $r_{i,done}=20$  when the $i$th agent has exited the simulation, the outflow improves by about 12\% compared with the human baseline (an outflow of $1749.78$ or $1744.96$), and by using both fully selfish reward and the exit bonus the outflow improves by additional $1.5\%$ (an outflow of $1771.74$). 
An additional improvement of $1.3\%-1.6\%$ is achieved by mixing a small fraction of global reward with a large fraction of selfish reward, as well as an exit bonus (an outflow of $1791.07$ and $1796.76$ for  the 0.2- and 0.1-collaborative policies).
We note, however, that due to the high variance in performance during training, the RL policies didn't always achieve these results during the training process. 
Generally, when the collaboration weight $\eta_2$ is less than $0.5$ and there is an exit bonus, the trained policies can all increase the average speed and outflow with respect to the human baseline, without significantly lowering the inflow.

The distributed policy outperforms the human baseline, but achieves slightly lower outflows than the centralized one.
The difference between the top performing distributed policy and the centralized one is not statistically significant.

\section{Conclusion and Future Work}
\label{sec:conclusions}
In this work we investigate reinforcement learning for traffic control in open networks. 
We show that the previously used metric of average speed is an insufficient measurement for open network traffic efficiency, since simulated inflows can be manipulated by the agents to achieve higher average speed. To address this deficiency, we propose the outflows of the network as a more reliable metric for evaluating traffic efficiency in open networks. 
We further show that by using the outflows as a reward function, our RL algorithm can generate a driving policy which is superior to a policy generated by the state-of-the art reward function in terms of both outflows, and average speed in a small open network.

After showing that existing methods cannot improve traffic efficiency in a large, realistic, open network (highway I-696 in Michigan, USA), we developed a modular transfer learning approach that applies the policy learned in a small network to a window surrounding a junction in the large network. Our results indicate that the modular approach achieves better outflows than both human-driven traffic, and a policy trained from scratch on the full network. On top of the improved traffic efficiency, a key advantage of the transfer learning approach is that it requires much less training time than a policy trained on the entire network, achieving an 80\% training time reduction in our setup.

Finally, we show for the first time that a distributed multiagent RL policy can improve traffic efficiency in a small open network, while relying on local knowledge sensed by the vehicles' internal sensors. This setting is more realistic than the centralized approach which relies on the combined information sensed by all the AVs.

An interesting avenue for expanding this research in future work is to try scaling the modular approach to more than one window. Similarly, another interesting direction is to try scaling the distributed approach to larger, more realistic scenarios.

\section{Acknowledgments}
This work has taken place in the Learning Agents Research
Group (LARG) at UT Austin.  LARG research is supported in part by NSF
(CPS-1739964, IIS-1724157, NRI-1925082), ONR (N00014-18-2243), FLI
(RFP2-000), ARO (W911NF-19-2-0333), DARPA, Lockheed Martin, GM, and
Bosch.  Peter Stone serves as the Executive Director of Sony AI
America and receives financial compensation for this work.  The terms
of this arrangement have been reviewed and approved by the University
of Texas at Austin in accordance with its policy on objectivity in
research.

\bibliographystyle{ACM-Reference-Format} 
\balance
\bibliography{references}  
\section*{Appendix}
\subsection*{A Hyper-parameters}
\subsubsection*{A.1 Centralized Agent Training}
\begin{table*}[h!]
  \centering
  \caption{Hyper-Parameters for Training Centralized Agents from Scratch}
  \label{hyperparameter1}
  \begin{tabular}{c|c}
     \specialrule{.1em}{.05em}{.05em}
     \textbf{Parameter} & \textbf{Value}\\
     \specialrule{.1em}{.05em}{.05em}
     Algorithm & Proximal Policy Optimization (PPO)\\
     \hline Horizon& 2000\\
     \hline Simulation Time Stepsize & 0.5 \\
     \hline Optimizer & Stochastic Gradient Descent\\
     \hline Learning Rate & $5\times10^{-4}$ \\
     \hline Discount Factor ($\gamma$) & 0.99\\
     \hline GAE Lambda ($\lambda$) & 0.97\\
     \hline Actor Critic & True \\
     \hline Value Function Clip Parameter & $10^6$\\
     \hline Number of SGD Update per Iteration & 10 \\
     \hline Model hiddens & [100,50,25] \\
     \hline Clip Parameter & 0.3 \\
     \hline Entropy Coefficient & 0 \\
     \hline Sgd Minibatch size & 128 \\
     \hline Train Batch Size & 40000\\
     \hline Value Function Share Layers & False \\
     \hline KL Coefficient & 0.2 \\
     \hline KL Target & 0.01 \\
     \hline Max Acceleration & 2.6\\
     \hline Max Deceleration & 4.5\\
     \hline Training Iterations & 500\\
     \hline Number of Rollouts per Iteration & 20\\
     \hline
     \specialrule{.1em}{.05em}{.05em}
\end{tabular}
\end{table*}
The hyper-parameter used to obtain the reward comparison results in centralized method on Simple Merge (Table \ref{tab:t1}) and on I-696 Merge (Table \ref{tab:3}) is as follows in Table \ref{hyperparameter1}

\subsubsection*{A.2 Distributed Agent Training}
\begin{table*}[h!]
  \centering
  \caption{Hyper-Parameters for Training Distributed Agents from Scratch}
  \label{hyperparameter2}
  \begin{tabular}{c|c}
     \specialrule{.1em}{.05em}{.05em}
     \textbf{Parameter} & \textbf{Value}\\
     \specialrule{.1em}{.05em}{.05em}
     Algorithm & Proximal Policy Optimization (PPO)\\
     \hline Horizon& 2000\\
     \hline Simulation Time Stepsize & 0.5 \\
     \hline Optimizer & Stochastic Gradient Descent\\
     \hline Learning Rate & piece-wise linearly decreasing starting from  $5\times10^{-4}$ (From scratch) \\
     \hline Discount Factor ($\gamma$) & 0.998\\
     \hline GAE Lambda ($\lambda$) & 0.95\\
     \hline Actor Critic & True \\
     \hline Value Function Clip Parameter & $10^8$\\
     \hline Number of SGD Update per Iteration & 10 \\
     \hline Model hiddens & [100,50,25] \\
     \hline Clip Parameter & 0.2 \\
     \hline Entropy Coefficient & $10^{-3}$ \\
     \hline Sgd Minibatch size & 4096 \\
     \hline Train Batch Size & 60000\\
     \hline Value Function Share Layers & True \\
     \hline Value Loss Coefficient & 0.5 \\
     \hline KL Coefficient & 0.01 \\
     \hline KL Target & 0.01 \\
     \hline Max Acceleration & 2.6\\
     \hline Max Deceleration & 4.5\\
     \hline Training Iterations & 500\\
     \hline Number of Rollouts per Iteration & 30\\
     \hline
     \specialrule{.1em}{.05em}{.05em}
\end{tabular}
\end{table*}
The hyper-parameter used to obtain the distributed feature augmentation results and reward hyper-parameter tuning on Simple Merge (Table \ref{tab:ablation} and Table \ref{tab:distributed_reward}) is as follows in Table \ref{hyperparameter2}
\subsubsection*{A.3 Human Vehicles}
\begin{table*}[h!]
  \centering
  \caption{Hyper-Parameters for Training Distributed Agents from Scratch}
  \label{hyperparameter3}
  \begin{tabular}{c|c}
     \specialrule{.1em}{.05em}{.05em}
     \textbf{Parameter} & \textbf{Value}\\
     \specialrule{.1em}{.05em}{.05em}
     \hline Controller & Sumo Default Controller(IDM) \\
     \hline Max Acceleration & 2.6\\
     \hline Max Deceleration & 4.5\\
     \hline Expected Time Headway & 1 second\\
     \hline
     \specialrule{.1em}{.05em}{.05em}
\end{tabular}
\end{table*}


\end{document}